\documentclass[sigconf]{acmart}

\usepackage[disable]{todonotes}

\usepackage{subcaption}
\AtBeginDocument{%
  \providecommand\BibTeX{{%
    \normalfont B\kern-0.5em{\scshape i\kern-0.25em b}\kern-0.8em\TeX}}}

\copyrightyear{2021}
\acmYear{2021}
\setcopyright{acmcopyright}\acmConference[GECCO '21]{2021 Genetic and
Evolutionary Computation Conference}{July 10--14, 2021}{Lille, France}
\acmBooktitle{2021 Genetic and Evolutionary Computation Conference (GECCO '21),
July 10--14, 2021, Lille, France}
\acmPrice{15.00}
\acmDOI{10.1145/3449639.3459317}
\acmISBN{978-1-4503-8350-9/21/07}

\begin{document}

\title{Evolving and Merging Hebbian Learning Rules: Increasing Generalization by Decreasing the Number of Rules}

\author{Joachim Winther Pedersen}
\email{jwin@itu.dk}
\affiliation{%
  \institution{IT University of Copenhagen}
  \streetaddress{2300 Copenhagen, Denmark}
  \city{Copenhagen}
  \country{Denmark}
  \postcode{43017-6221}
}

\author{Sebastian Risi}
\email{sebr@itu.dk}
\affiliation{%
  \institution{IT University of Copenhagen}
  \streetaddress{}
  \city{Copenhagen}
  \country{Denmark}}
\email{}


\begin{abstract}

Generalization to out-of-distribution (OOD) circumstances after training remains a challenge for artificial agents. To improve the robustness displayed by plastic Hebbian neural networks, we evolve a set of Hebbian learning rules, where multiple connections are assigned to a single rule. Inspired by the biological phenomenon of the genomic bottleneck, we show that by allowing multiple connections in the network to share the same local learning rule, it is possible to drastically reduce the number of trainable parameters, while obtaining a more robust agent. During evolution, by iteratively using simple K-Means clustering to combine rules, our \emph{Evolve \& Merge} approach is able to reduce the number of trainable parameters from 61,440 to 1,920, while at the same time improving robustness, all without increasing the number of generations used. While optimization of the agents is done on a standard quadruped robot morphology, we evaluate the agents’ performances on slight morphology modifications in a total of 30 unseen morphologies. Our results add to the discussion on generalization, overfitting and OOD adaptation. To create agents  that can adapt to a wider array of unexpected situations, Hebbian learning combined with a regularising ``genomic bottleneck'' could be a promising research direction. 
\end{abstract}



\maketitle

\section{Introduction}

The inability to adapt to out-of-distribution (OOD) situations makes artificial agents less useful and despite many advances in the field of reinforcement learning and evolutionary strategies, making adaptable agents remains a challenge \cite{zhang2018dissection, zhang2018study, zhao2019investigating,justesen2018illuminating}. If we want agents to be deployed in complex environments, we cannot expect to be able to train them on all possible situations beforehand.

While artificial neural networks (ANNs) were originally inspired by the brain, they differ from biological neural networks on several key points. One crucial difference is that most ANNs have their connection strengths trained during a training phase, after which they stay fixed forever. Biological neural networks on the other hand undergo changes throughout their lifetimes. The framework of Evolved Plastic Artificial Neural Networks (EPANNs) \cite{soltoggio2018born} aims to incorporate the plasticity – and thereby the adaptability – of biological neural networks into artificial agents. The brain architectures found in biological agents have been shaped by evolution throughout millions of years \cite{breedlove2013biological}. The specific wiring of the synapses within each individual brain is a result of learning to adapt to sensory input, faced within the lifetime of the individual \cite{power2017neural, stiles2000neural, greenough2013induction}. 

\begin{figure}
\includegraphics[scale=0.25]{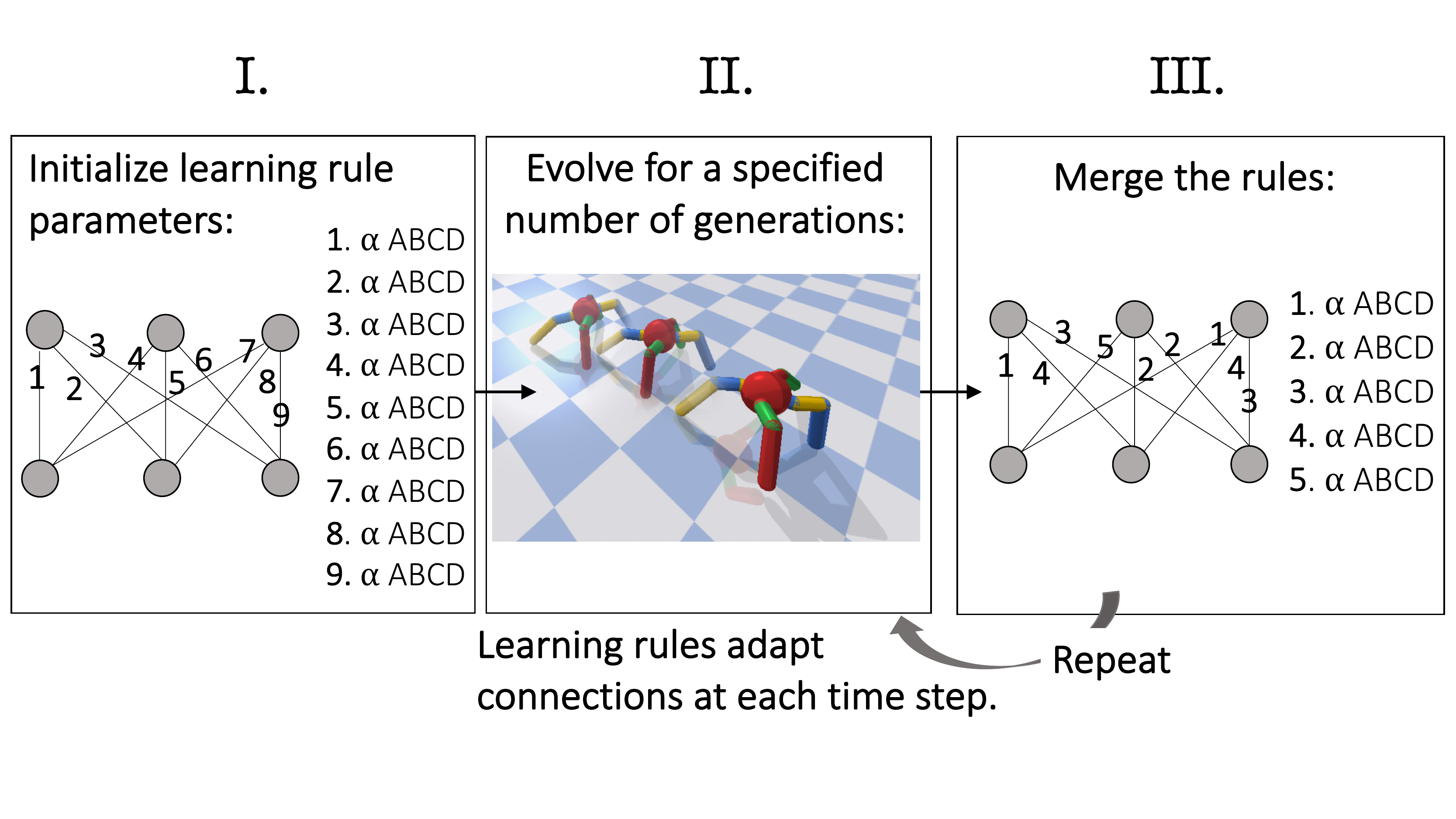}
\caption{Evolve and Merge Approach. \normalfont First, one learning rule is randomly initialized for each connection in the network. Then, the learning rule parameters are evolved for a predetermined number of generations. Subsequently, the rules that have similar parameters are  merged to a single rule. The new reduced rule set is then evolved further. This process is repeated and continues until the maximum number of generations allowed has been reached.}
\centering
\label{fig:overview}
\end{figure}

Here, we build upon the approach introduced in Najarro and Risi~\cite{najarro2020}, where parameters of local plasticity rules - not the connections of the network - were evolved. With this approach, ANNs with plastic connections showed better performances than static ANNs when faced with changes in robot morphology not seen during training.

However, in contrast to nature, in which  genomes encode an extremely compressed blueprint of a nervous system, the approach by Najarro and Risi~\cite{najarro2020} required each connection in the network to have its own learning rule, significantly increasing the amount of trainable parameters. 

The amount of information it takes to specify the wiring of a sophisticated brain is far greater than the information stored in the genome \cite{breedlove2013biological}. Instead of storing a specific configuration of synapses, the genome is thought to encode a much smaller number of rules that govern how the wiring should change throughout the lifetime of the individual \cite{zador2019critique}. In this manner, learning and evolution are intertwined: evolution shapes the rules that in turn shape learning \cite{hinton1996learning,price2003role, snell2013overview}. If the rules encoded by the genome do not allow the individual to learn useful behavior, then these rules will have to be adapted or go extinct.
The phenomenon that a large number of synapses has to be controlled by a small number of rules has been called the "genomic bottleneck", and it has been hypothesized that this bottleneck acts a regularizer that selects for generic rules likely to generalize well \cite{zador2019critique}.

Here, we aim to mimic the genomic bottleneck by limiting the number of rules to be much smaller than the number of connections in the neural networks, in the hope that we can evolve more robust agents.
The main insight in the novel approach introduced in this paper (Figure~\ref{fig:overview}) is that after having optimized one learning rule for each synapse in the network, it is possible with a simple clustering approach to drastically reduce the number of unique learning rules required to achieve good results. In fact, we show that starting from having one rule per connection in the ANN (12,288 rules) we can decrease this number of rules by 96.875 \% to have only 384 rules controlling all 12,288 connections. We further show that as the number of learning rules decreases, the robustness to unfamiliar morphologies tends to increase. In addition to be inspired by the compressed representations in genomes, the method proposed in this paper is related to the field of indirect encoding, which has a long history in artificial intelligence research \cite{gruau1996comparison, bentley1999three, stanley2003taxonomy, toneli2013}.

As we gradually decrease the number of learning rules, we end up with a set of rules, which have a smaller number of trainable parameters than there are connections in the network. We operationalize robustness as being able to perform well across an array of settings not seen during training. We compare the robustness of plastic ANNs to that of different static ANNs (described further in section \ref{subsection:experiments}): a plain static network with the same architecture as our plastic networks, a smaller static network, and static networks where noise is applied to their inputs during optimization. The plastic networks outperform the plain static networks of different sizes in terms of robustness, and perform at the same level as the best configurations of static networks with noisy inputs while requiring a significantly lower number of trainable parameters.

In the future it will be interesting to further increase the expressivity of the evolved rules, potentially allowing an even greater genomic compression with increased generalisation to robots with drastically different morphologies.  

\section{Related Work}
This section reviews related work on Hebbian learning and indirectly encoding plasticity in evolving neural networks.

\subsection{Hebbian Plasticity}\label{subsection:heb}
In neuroscience, Hebbian theory proposes that if a group of neurons are activated close to each other in time, the neurons can increase their efficacy in activating each other at a later point \cite{holscher2008could}. In this way, the full group could thus become active, even if just a part of the group was initially activated by external stimuli \cite{lansner2009associative}. This type of plasticity constitutes a local learning rule, where the change in connection only depends on information local to the two connected neurons, such as their activation. In this way, recurrent connections within a layer of neurons endow a network with the ability to ``complete patterns''\cite{hunsaker2013operation}: if a stimulus at an earlier time has activated a certain group of neurons, also called a cell assembly, the network might be able to respond with the activation of this same cell assembly, if it is faced with an incomplete or noisy presentation of this stimulus \cite{hoffmann2009face}. These cell assemblies have been proposed to be the basic units for thoughts and cognition \cite{buzsaki2010neural, pruszynski2019language, saxena2019towards}, and their existence was first suggested by Hebb \cite{pulvermuller2014thinking}.

Several past studies have focused on finding ways to use plasticity inspired by Hebbian plasticity in artificial neural networks. The motivation for this has for some studies been to study network plasticity in computational models \cite{song2000competitive, abbott2000synaptic}, and for other studies the aim has been to better enable models to generalize. For example, Soltoggio et al. \cite{soltoggio2008evolutionary} evolve modulatory neurons that when activated will alter the connection strengths between other neurons. With this approach, they are able to solve the T-Maze problem, where rewards are not stationary. In Orchard and Wang~\cite{orchard2016evolution}, linear and non-linear learning rules are evolved to adapt to a simple foraging task. Yaman et al. \cite{yaman2019learning} use genetic algorithms to optimize delayed synaptic plasticity that can learn from distal rewards. Common to these examples are that learning rules that update neural connections have access to a reward signal during the lifetime of the agent. While this might be a useful future addition to our approach, here the learning rules only have access to neural activations, requiring adaptation to only rely on the robot's proprioceptors. 

\subsection{Indirect Encoding with Plasticity}
One of the goals of indirect encoding approaches is to be able to represent a full solution, like a large neural network, in a compressed manner \cite{bentley1999three, gruau1996comparison, stanley2003taxonomy}. 
Indirect encoding schemes come in many variations. Examples include letting smaller networks determine the connection strengths of a larger network \cite{ha2016hypernetworks, risi2012enhanced, risi2011enhancing, carvelli2020evolving}, or representing the connections of a neural network by Fourier coefficients \cite{koutnik2010evolving, gomez2012compressed}. 
Of particular relevance to our work are methods that use plasticity rules to make indirect encodings. 

Early approaches include that of Chalmers \cite{chalmers1991evolution}, where a single parameterized learning rule with 10 parameters was evolved to allow a feedforward network to do simple input-output associations. More recently, approaches such as \emph{adaptive HyperNEAT} \cite{risi2010indirectly} have been deployed to exploit the geometry of a neural network to produce patterns of learning rules.  The \emph{HyperNEAT} approach has previously been used for improving a controller's ability to control robot morphologies outside of what was experienced during training \cite{risi2013confronting}. However, this method depended on explicit information about the morphologies, whereas the method used in the current paper does not rely on any such information.  Indirectly encoding plasticity has shown to improve a network's learning abilities, with networks that are more regular showing improved performance \cite{toneli2013}. These earlier results point at a deep connection between plasticity and indirect encodings, which has so far received little attention. 

\section{Approach}

The approach introduced in this paper evolves a set of local learning rules, where the number of rules is ultimately much smaller than the number of connections. In contrast to other indirect encoding methods, instead of starting with a small rule set, the \emph{Evolve \& Merge} approach starts with a large amount of trainable parameters compared to the number of trainable parameters in the ANN and only over the course of the evolution end up with a smaller number of trainable parameters by merging rules that have evolved to be very similar (Figure~\ref{fig:overview}).

For learning rules, we use a parameterized abstraction of Hebbian learning. The so-called "ABCD" rule, which has been used several times in the past as a proxy for Hebbian learning \cite{najarro2020, niv2002evolution, risi2010evolving, orchard2016evolution}, updates the connection between two neurons in the following manner:
\begin{equation}
  \Delta w_{ij} = \alpha(Ao_{i}o_{j} + Bo_{i} + Co_{j} + D)
\end{equation}
where $w_{ij}$ is the connection strength between the neurons, $o_{i}$ and $o_{j}$ are the activity levels of the two connected neurons, $\alpha$ is a learned learning rate, and $A$, $B$, $C$, and $D$ are learned constants.

Instead of directly optimizing connection strengths in the plastic network, we only optimize parameters of the ABCD learning rules, which in turn continually adapts the network’s connections throughout the lifetime of the agent.
The parameters of the learning rules are randomly initialized before evolution following a normal distribution, $~\mathcal{N}(0,0.1)$, and are optimized at the end of each generation. 
In the beginning of each new episode, the connection strengths of a plastic network are randomly initialized, drawn from a uniform distribution, $~Unif(-0.1, 0.1)$. At each time step of the episode, after an action has been taken by the agent, each connection strength is changed according to the learning rule that it is assigned to. Below, we show training and performance of models for which a unique learning rule is optimized for each connection in the network, as well as models where multiple connections share the same learning rule.

\subsection{Environment}
We train and evaluate our models on how well they can control a simulated robot in the AntBullet environment \cite{benelot2018}. Here, the task is to train a three-dimensional four-legged robot to walk as efficiently as possible in a certain direction.

The neural network controlling the robot receives as input a vector of length 28, in which the elements correspond to the position and velocity of the robot, as well as angles and angular velocities of the robot’s joints.  To control the robot’s movements, the output of the neural network is a torque for each of the eight joints of the robot, resulting in a vector of eight elements.

\begin{figure}
\includegraphics[scale=0.23]{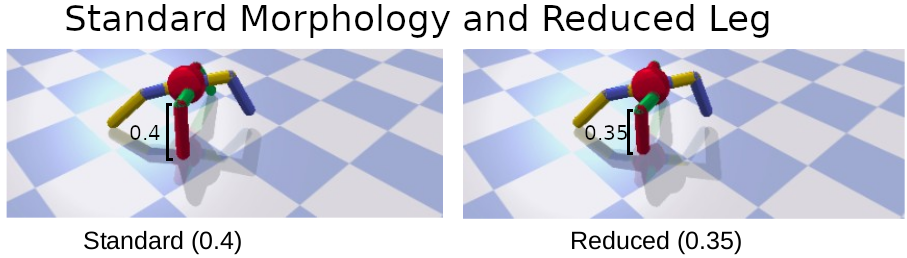}
\caption{Robot Environment. \normalfont  Right: The standard morphology that the models were optimized for. Left: An example of the most severe leg reduction done to a leg to test robustness. The lower part of the leg was reduced from 0.4 to 0.35. In this figure, the reduction was done to the red part of the leg in the foreground of the image. }

\centering
\label{fig:environment}
\end{figure}

Previous research has mostly focused on a robot’s ability to cope with catastrophic damages to a leg \cite{cully2015robots, colas2020scaling, najarro2020}. Here we take a different approach, where legs are modified, but still usable (Figure~\ref{fig:environment}).  For evaluation, we create slight variations on the standard morphology by shrinking the lower part of the robot’s legs (the “ankles” in the robot’s xml file). The standard length of an “ankle” is 0.4. We evaluate five different reductions, reducing the length of an “ankle” to 0.39, 0.38, 0.37, 0.36, and 0.35, respectively. One by one, each of the four legs have their ankle reduced to each of these five lengths. In addition, we evaluate on a setting where both front legs are reduced at the same time, as well as a setting where the front left leg and the back right leg are reduced at the same time. We thus have five different reductions on six different leg combinations, amounting to 30 different morphology variations. This way we can more precisely determine how much variation from the original setting the models are able to cope with.
When evaluating on a varied morphology, we adapt the reward to solely reflect the distance traveled in the correct direction, so that no points are given for just "staying alive".

\subsection{Evolution Strategy}
All models are optimized using Evolution Strategy (ES) \cite{salimans2017evolution}. In static networks connections are evolved directly and in plastic networks only the learning rule parameters are evolved. We use an off-the-shelf implementation of ES \cite{ha2017evolving} with its default hyperparameters, except that we set weight decay to zero (see Table ~\ref{tab:hyper} for complete hyperparameter configurations). This implementation uses mirrored sampling, fitness ranking, and uses the Adam optimizer for optimization. In all runs, a population size of 500 is used, and optimization spans 1,600 generations.

\begin{table}
  \caption{Hyperparameters for ES}
  \label{tab:hyper}
  \begin{tabular}{cc}
    \toprule
    Parameter& Value\\
    \midrule
    Population Size & 500\\
    Learning Rate & 0.1\\
    Learning Rate Decay & 0.9999\\
    Learning Rate Limit& 0.001\\
    Sigma & 0.1\\
    Sigma Decay& 0.999\\
    Sigma Limit & 0.01\\
    Weight Decay & 0\\

  \bottomrule
\end{tabular}
\end{table}

\

\subsection{Merging of Rules}
\begin{figure}
\includegraphics[scale=0.116]{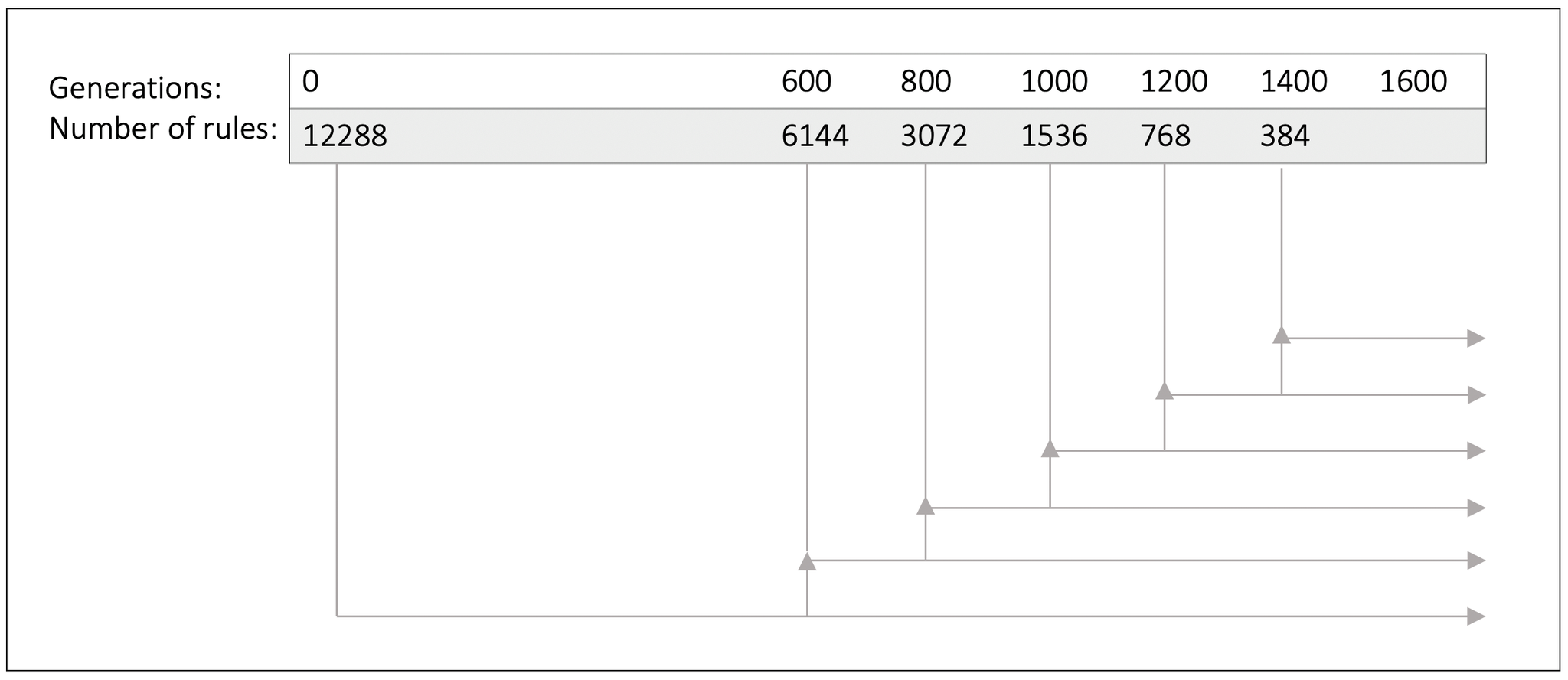}
\caption{Rule Merging. \normalfont The number of rules is iteratively halved using K-Means clustering. Every reduced rule set is optimized until the time limit of 1,600 generations so that fair comparisons between models can be made.}
\centering
\label{fig:pruning}
\end{figure}

In order to produce rule sets with fewer rules than the number of connections in the network, we use the K-Means clustering algorithm \cite{scikit-learn} to gradually merge the learning rules throughout training. We first initialize a network with one learning rule for each synapse (12,288 different learning rules) and optimize these learning rules for 600 generations. We then half the number of learning rules by using the K-Means algorithm to find 6,144 cluster centers among the 12,288 rules. 
The new learning rule of a given synapse will simply be the cluster center of the learning rule, to which it was previously assigned. The newly found smaller set of learning rules is then optimized for 200 generations, before K-Means clustering is used to half the number of rules again. The number of generations between merging of the rules as well as how many cluster centers to reduce the rule set to are hyperparamters that were determined to work well in preliminary experiments, and the choices mainly depend on how much total training time is permitted. This process is repeated until we have optimized for 1,600 generations altogether.  In order to be able to make fair comparisons of the different reduced rule sets, we also optimize each individual reduced set until it has been optimized for 1,600 generations (see Figure~\ref{fig:pruning}).

\subsection{Experiments}
\label{subsection:experiments}

We compare a total of 12 different models in terms of their ability to learn and their robustness (see Table ~\ref{tab:scores} and Figure~\ref{fig:box_mean}, \ref{fig:box_worst}). Common to all of them is that they are feedforward networks, they have two hidden layers, they have no biases, and the hyperbolic tangent function is used for all activations. 

All networks, have 128 neurons in the first hidden layer, and 64 neurons in the second. The \textbf{plain static model} is a static neural network without any Hebbian learning. We also optimize and evaluate a \textbf{smaller static model}, in which the hidden layers are size 32 and 16 (1,536 parameters).  Two other static network models were trained in a setting where \textbf{noise} was applied to the input throughout optimization. If the models can perform well despite noisy inputs, they might also be more robust to morphology changes despite being static, and therefore we create these models as additional baselines to compare our approach to. At each time step during optimization, a vector of the same size as the input with elements drawn from a normal distribution was created and then added to the input element-wise. For one model, the distribution was $\mathcal{N}(0, 0.05)$, the other had a distribution of $\mathcal{N}(0, 0.1)$. In the figures below, these models are named after the amount of noise their inputs received during optimization.

\noindent The \textbf{ABC} model type was trained with an incomplete ABCD rule: the learning rate and the D parameter were omitted, so only the activity dependent terms were left.  For all other plastic network models, the rules consisted of five parameters (A,B,C,D and the learning rate).  The model called $\boldsymbol{\alpha} \textbf{ABCD}$ was optimized with a rule for each connection throughout evolution, and the number of rules was never reduced. This is the same method as was introduced by Najarro and Risi \cite{najarro2020} \todo{added line to clarify that this was the original approach}. The model called "$\textbf{500 rules from start}$" was initialized with just 500 rules, and this number remained unchanged. Each connection of the ANN was randomly assigned to one of the 500 rules at initialization. In the figures summarizing the results (Figures ~\ref{fig:mean_score},\ref{fig:orig_scores}, \ref{fig:box_mean}, \ref{fig:box_worst}), the rest of the plastic models are simply named after the number of rules they have in their rule sets.

\section{Results}
\begin{figure}
\includegraphics[scale=0.18]{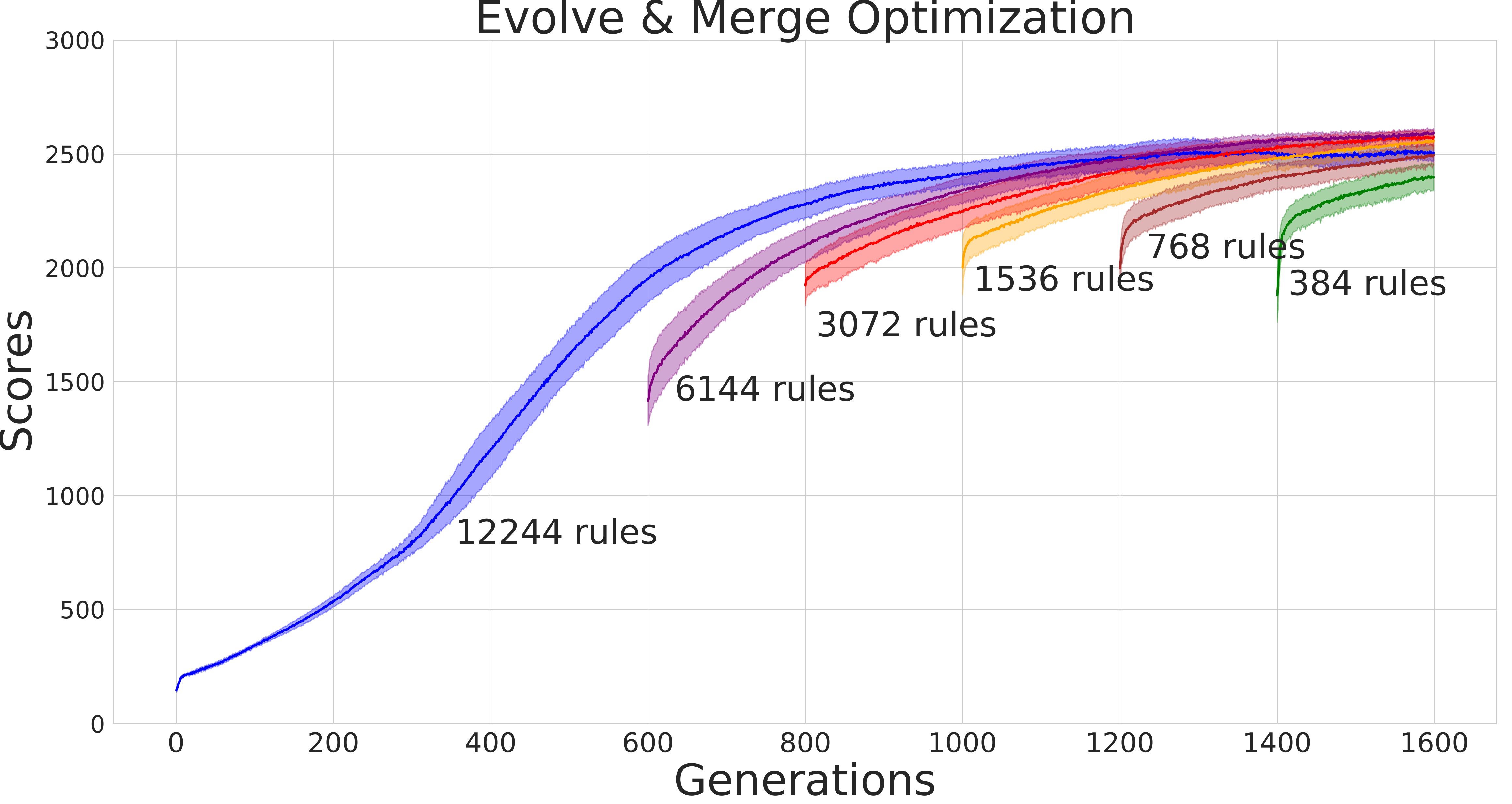}
\caption{Evolve \& Merge Training Results. \normalfont Training curves of five independent evolution runs. The solid lines reflect the average population mean for each run of each model. The filled areas are the standard deviations of the models' population means. Immediately after a merging of rules has occurred, the newly reduced rule set is set back a bit in its performance compared to before merging, but performance quickly recovers and improves. Without adding to the total optimization time of 1600 generations, a rule set of just 384 different rules reaches a comparable population mean score as a rule set with 12244 rules.}
\centering
\label{fig:evolve_merge_training}
\end{figure}

\begin{figure}
\includegraphics[scale=0.19]{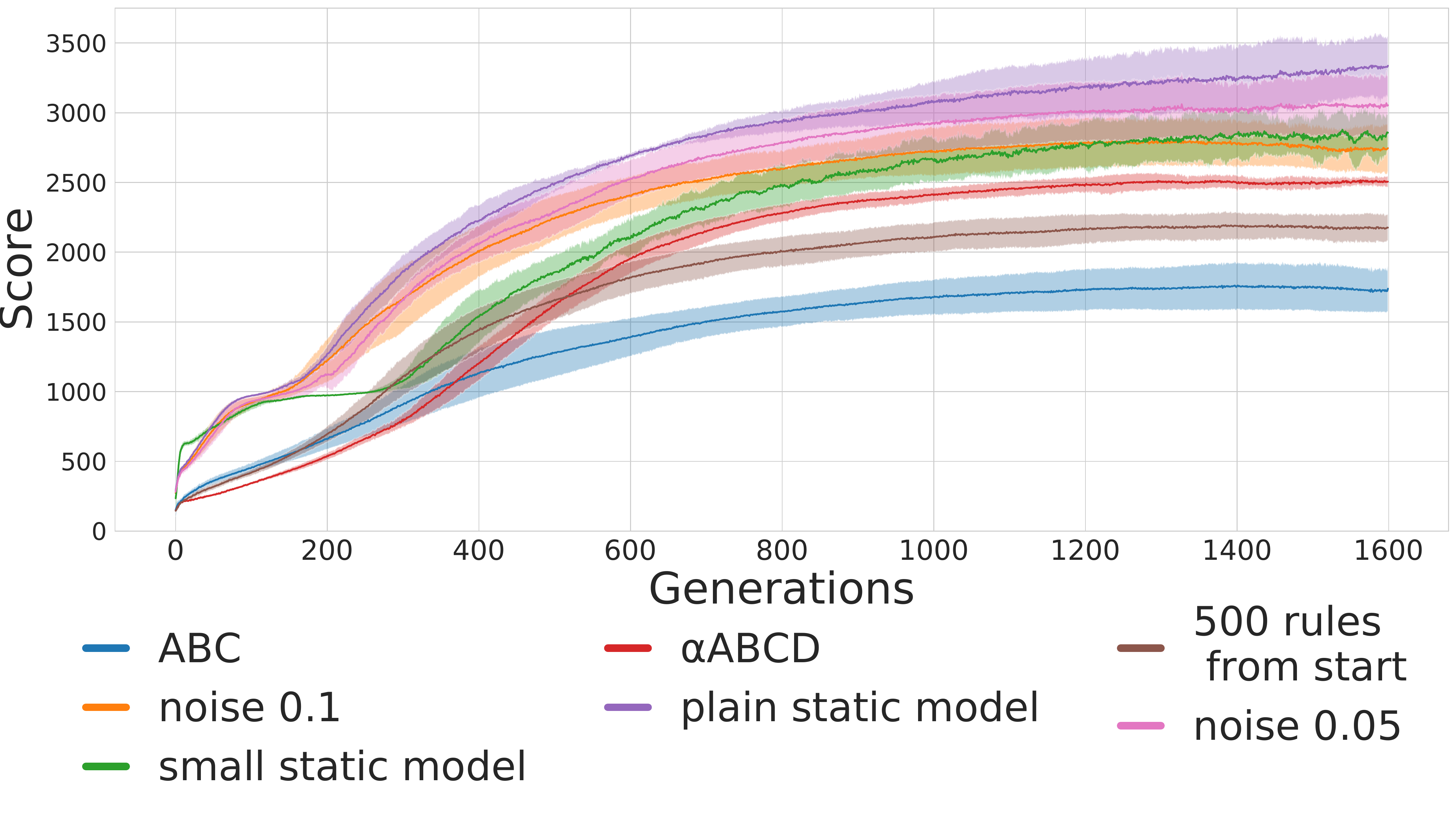}
\caption{Training Results. \normalfont For all models, we ran five independent evolutionary runs. The solid lines reflect the average population mean for a given model throughout evolution. The filled areas are the standard deviations of the models' population means. While the static models end up with better scores, all models, except 'ABC', are able to obtain reasonably well-performing populations within 1600 generations.} 
\centering
\label{fig:train_curves}
\end{figure}

After optimizing for 1,600 generations, we see that the static networks tend to perform much better on the standard morphology that the models are optimized on. Figure~\ref{fig:evolve_merge_training} presents the training curves of the each of the reduced rule sets, and Figure~\ref{fig:train_curves} shows the training curves of the rest of the models. For comparisons with different reinforcement learning algorithms in the standard AntBullet environment see Pardo  \cite{pardo2020tonic} for baselines where, e.g., Proximal Policy Optimization (PPO) achieves a score of around 3100, Deep Deterministic Policy Gradient (DDPG) scores around 2500, and Advantage Actor-Critic (A2C) scores around 1800.

\begin{table}

  \begin{tabular}{lclc}
    \toprule
    Model Name& Num. Params.&Orig. Score&Novel Score\\
    \midrule
    plain static model & 12,288& \textbf{3,513}$\pm$221 &$2,095\pm$442\\
    small static model & 1,536& 3,040$\pm$170 &923$\pm$387\\
    noisy (0.05) & 12,288&3,283$\pm$213 &2,172$\pm$395\\
    noisy (0.1) & 12,288& 3,020$\pm$126 &2,256$\pm$226\\
    500 from start & 2,500& 2,185$\pm$92 &1,931$\pm$90\\
    ABC & 36,864& 1,731$\pm$158 &1,462$\pm$206\\
    $\alpha$ ABCD & 61,440& 2,528$\pm$52 &2,173$\pm$126\\
    $6,144$ rules & 30,720& 2,649$\pm$13&2,259$\pm$100\\
    $3,072$ rules & 15,360& 2,631$\pm$21 &2,244$\pm$123\\
    $1,536$ rules & 7,680& 2,631$\pm$27 &\textbf{2,323}$\pm$90\\
    $768$ rules& 3,840& 2,580$\pm$33 &2,286$\pm$157\\
    $384$ rules& \textbf{1,920}& 2,516$\pm$44 &2,267$\pm$186\\
  \bottomrule
\end{tabular}
\caption{Parameter counts and scores after optimization. \normalfont Orig. Scores are scores under original morphology settings, and Novel Scores are under the unseen altered settings. Scores reflect the means of 5 models of each model type. A model's score is its mean performance over 100 episodes. Standard deviations are provided next to each mean. The plain static network has the highest score in the original settings, but has a massive decrease in score in the novel settings. All reduced rule sets have higher mean scores with smaller standard deviations in the novel settings, than any static network.}
 \label{tab:scores}
\end{table}

\begin{figure}
\includegraphics[scale=0.18]{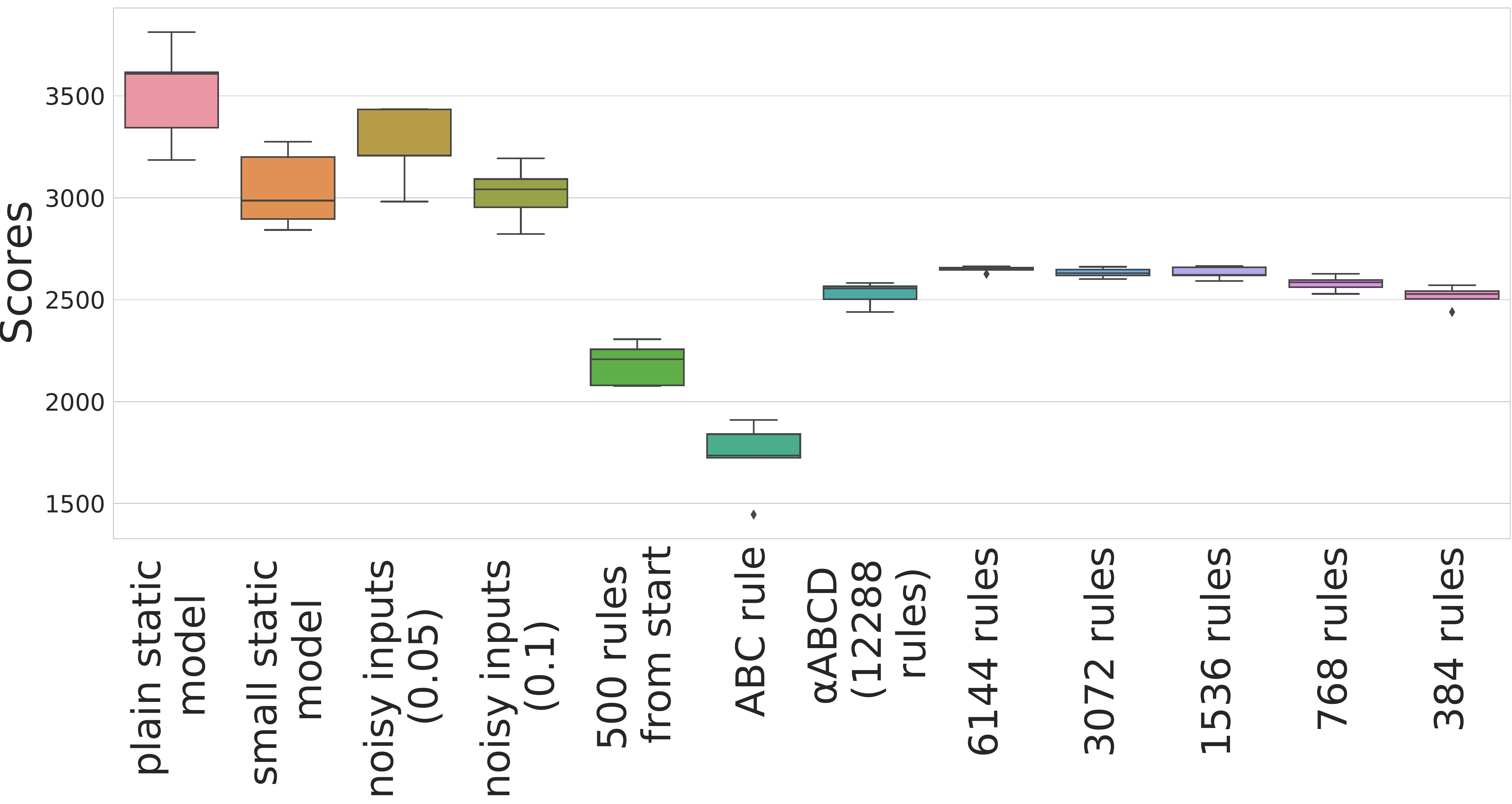}
\caption{Original Environment Scores. \normalfont Box plots for the scores of the optimized models in the original environment setting. For each model, the score is averaged over 100 independent episodes. As we optimized 5 models of each type, the box plots show the variation of the scores within a given model type. See Section ~\ref{subsection:experiments} for model descriptions. All static models have better scores than any plastic model in the original setting. The reduced rule sets see no decrease in performance compared to the model with a rule for each connection.}
\centering
\label{fig:orig_scores}
\end{figure}

\begin{figure}  
\includegraphics[scale=0.175]{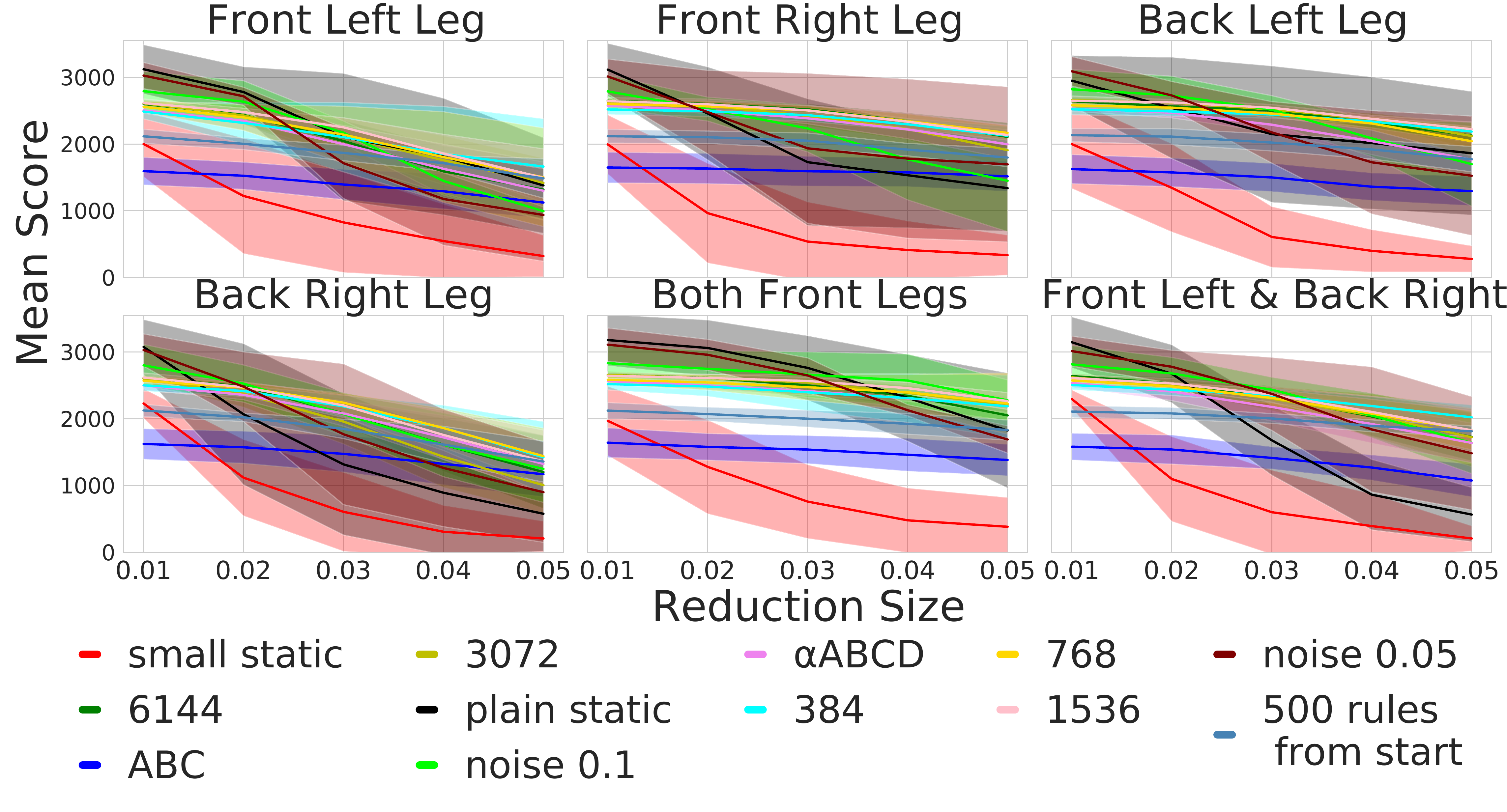}
\caption{Mean scores for all models in all novel environment settings. \normalfont As leg reductions increase (x-axis), the models tend to perform worse (y-axis). The mean scores are calculated over 100 episodes for each model. The static models have much more variability in their scores than the plastic ones. Further, static networks tend to start with a good score for the smallest reduction, but decrease rapidly as reductions become larger. The plastic networks, on the other hand, have much flatter performance curves.}
\label{fig:mean_score}
\end{figure}

\begin{figure*}
     \centering
     \label{fig:boxes}
     \begin{subfigure}[b]{0.49\textwidth}
         \centering
         \includegraphics[width=\textwidth]{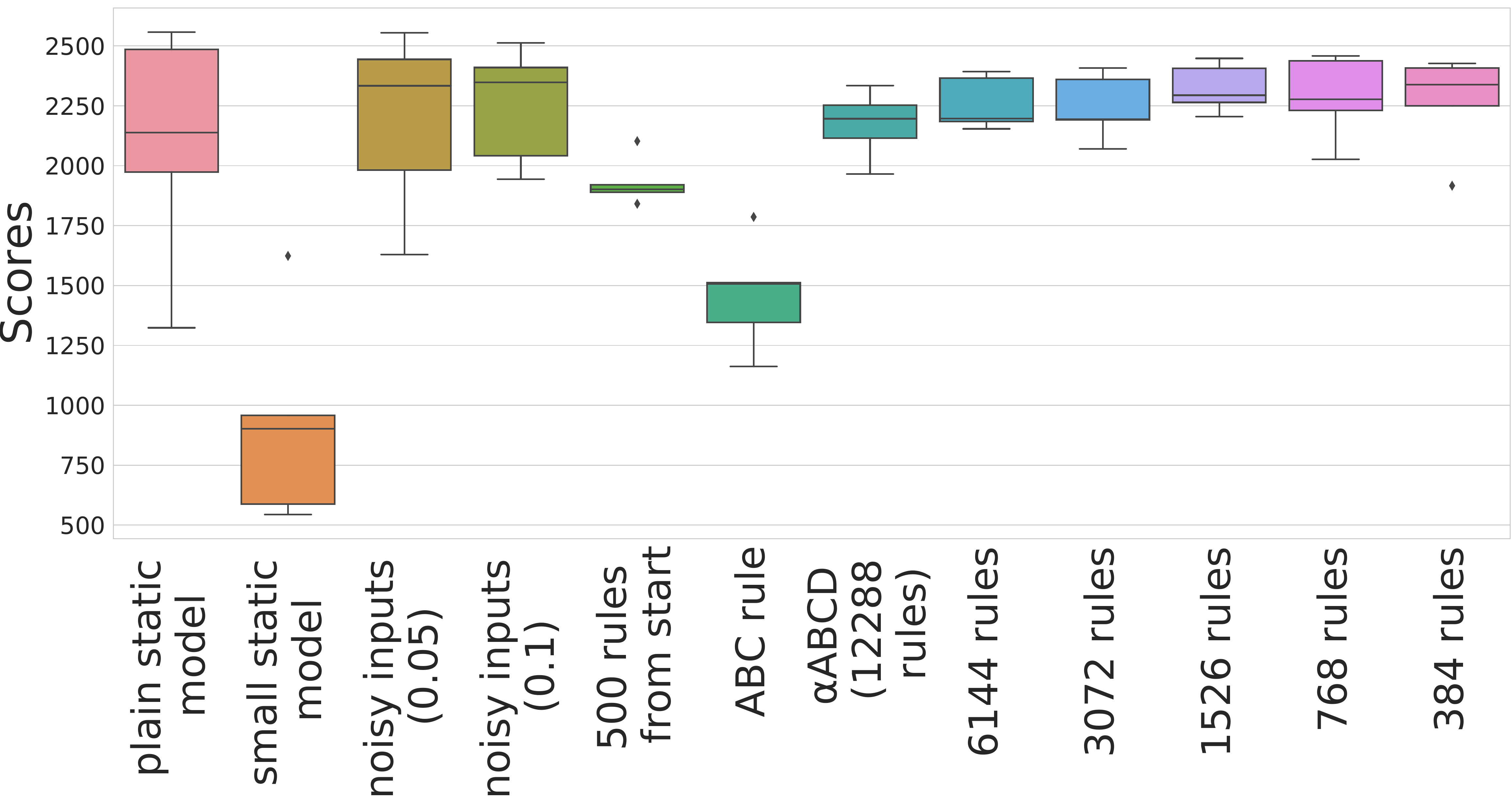}
         \caption{Mean Scores}
         \label{fig:box_mean}
     \end{subfigure}
          \begin{subfigure}[b]{0.49\textwidth}
         \centering
         \includegraphics[width=\textwidth]{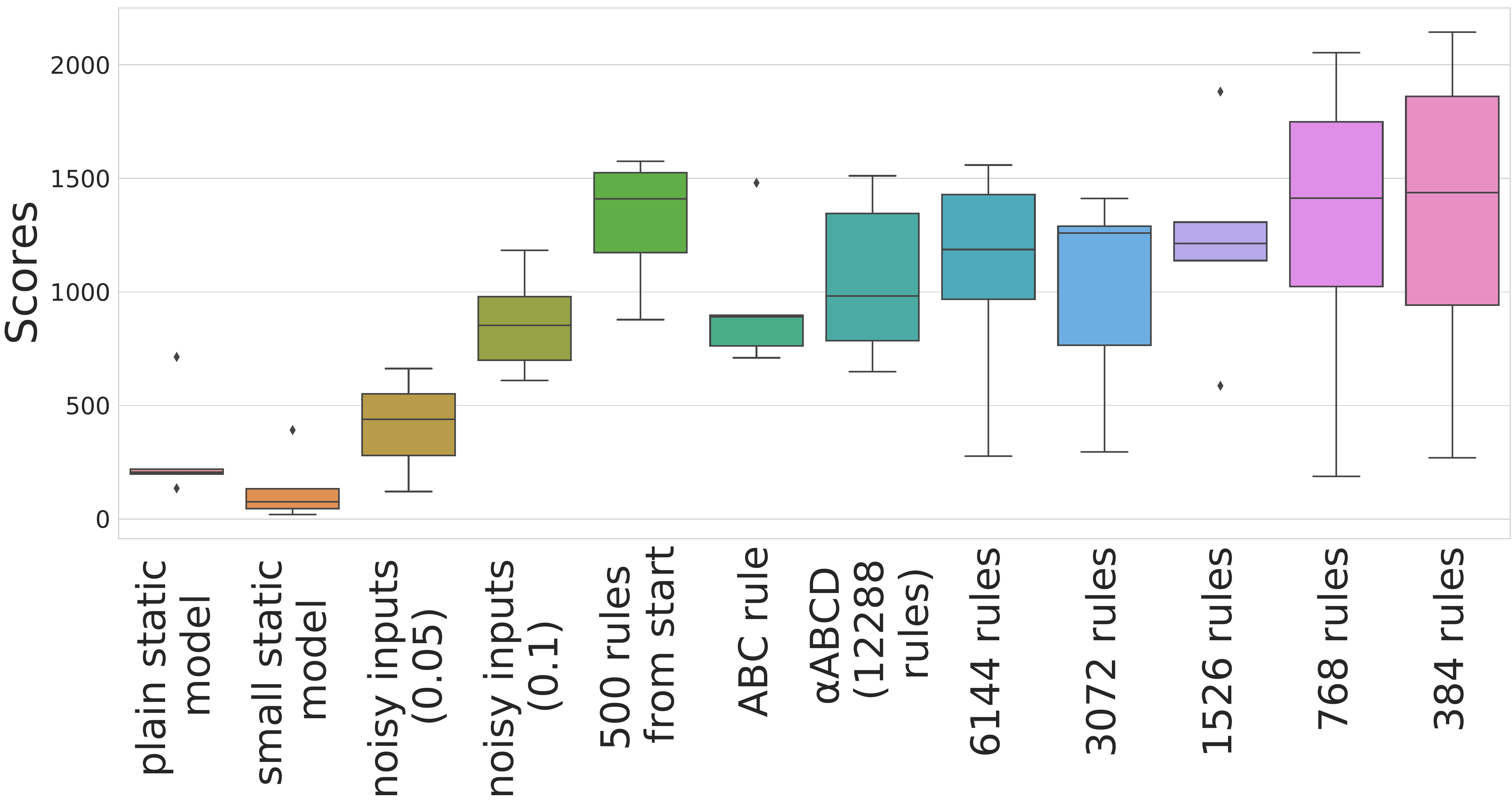}
        \caption{Worst Mean Scores}
         \label{fig:box_worst}
     \end{subfigure}
     \caption{Generalisation Performance. \normalfont Box plots for the (a) mean scores and (b) worst mean scores of the optimized models across all novel environment setting. For each model, the score is averaged over 100 independent episodes. The worst mean scores shown in (b) shows how bad a model is at its worst across 100 episodes. The worst scores of the plastic networks tend to be much better than the worst scores of the static one. This shows that plastic models are at less risk of getting a catastrophically bad score where the robot is only barely able to move at all. As in Figure~\ref{fig:orig_scores}, the box plots show the variation of the scores within a given model type. }
     
\end{figure*}

Performances after optimization are summarized in Table~\ref{tab:scores} and Figures ~\ref{fig:orig_scores},\ref{fig:mean_score}, \ref{fig:box_mean}, \ref{fig:box_worst}. 
The reduced rule sets generally end up with a better performance when evaluated the original settings compared with the case where the number of rules is equal to the number of connections throughout.
Looking at the box plots in Figure ~\ref{fig:box_mean}, we see that the all models with reduced rule sets (named by number of rules) have a higher mean performance across all novel settings than the model with a rule for each connection (named  $\boldsymbol{\alpha} \textbf{ABCD}$ in the figures). They also have a higher mean performance than the plain static model, which have a large variability in its performance across the its different evolution runs. The average performance tends to increase as the number of rules decreases.
The \textbf{ABC} model and the model that had only 500 rules from the beginning achieved training performances considerably lower than the rest of the models.
The static models optimized with noisy inputs obtain similar scores as the reduced rule sets. When the noise added to the input is drawn from $\mathcal{N}(0, 0.05)$, the best evolution run of the model is, like the static network, better than any of the runs with the reduced rule sets, but its worst run is far worse. When the noise is drawn from $\mathcal{N}(0, 0.1)$ the average performance is at the same level as the most reduced rule set and the top performance is a bit better.
Overall, while the ceiling of the performances seems to be lower for the plastic networks, the floor is tends to be much higher (except for the case of the ABC-only models). The latter point is especially visible if we look at the worst performances of the runs of each model, as is done in Figure~\ref{fig:box_worst}. Here we see that the model trained with 500 learning rules from the beginning of the training is the least likely to get catastrophically bad performances.

\subsection{Discovered Rules}
\begin{figure*} 
\includegraphics[scale=0.58]{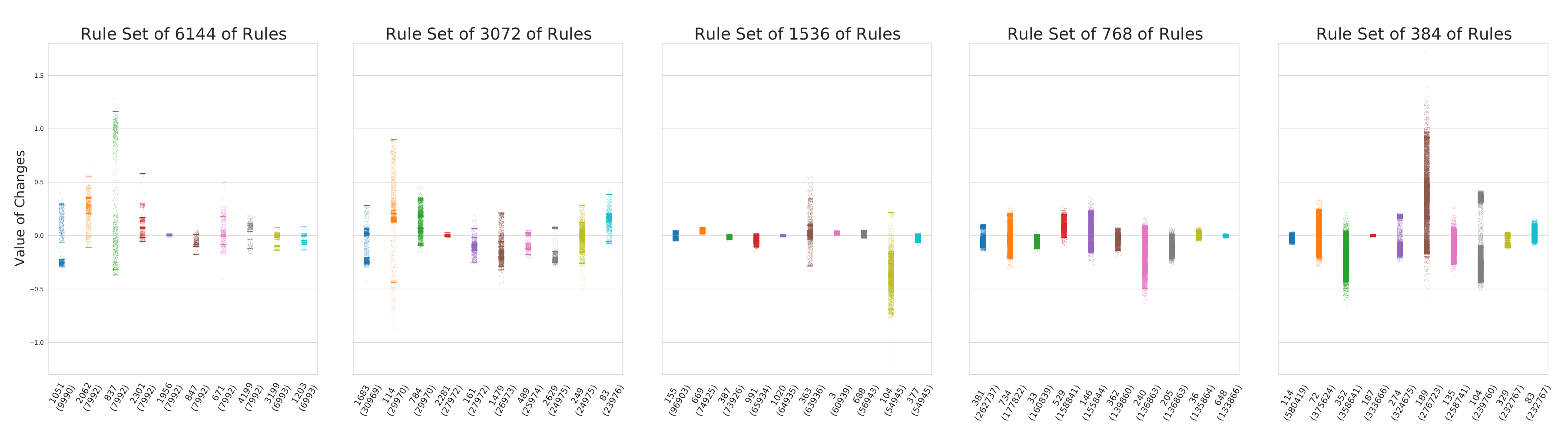}
\caption{Evolved Rules Examples. \normalfont Strip plots of values of connection weight changes of the top 10 most followed rules in an optimized version of each of the reduced rule sets. The names of each of the strips seen along the x-axis are the indices of the rules, and in parentheses are the number of updates that the rule made during an episode with the original environment settings. A strip indicates what updates was made by the rules during this episode.}
\centering
\label{fig:strip}
\end{figure*}


To get a visual intuition of what type of learning rules evolved, Figure~\ref{fig:strip} shows strip plots of the top 10 most followed rules in each of the reduced rule sets. From these, we can see that several different types of rules are found by evolution. First, one type of rule that is found in the top 10 of all the rule sets, have all its updates closely concentrated around zero (e.g., the rule indexed 1020 in the rule set of 1536 rules, or the rule indexed 114 in the set of 384 rules). Second, multiple rules provide relatively strong updates, both positive and negative, but which have a gap around zero (e.g., index 104 in the 384 set, or 837 in the 6144 set). Another apparent type of rule provides both positive and negative updates, but has no gap around zero (e.g., index 734 in the 768 set, or 248 in the 3072 set). Lastly, we see a type of rule, which has the vast majority of its updates to be of a specific sign (e.g, index 240 in set 768 (mostly negative), or 784 in set 3072 (mostly positive)). These observations suggest that many connections are destined to only receive very small updates throughout the episode, and remain largely unchanged compared to some of the stronger updates that other rules provide. Further, the fact that multiple of the rules that have the most connections assigned to them, can provide both positive and negative updates, confirms that two connections following the same rules might end up being updated very differently from each other. Fewer rules do therefore not necessarily make the neural network less expressive.

\section{Discussion}

In this paper, we build upon the results of Najarro and Risi~\cite{najarro2020} that showed that an increased robustness can be achieved by evolving plastic network with local Hebbian learning rules instead of evolving ANN connections directly. We show that the robustness can be enhanced even further with an \emph{Evolve \& Merge} approach: throughout optimization we iteratively use a clustering algorithm to merge similar rules, resulting in a smaller rule set at the end of the optimization process. While the plastic networks are not able to get scores as high as the highest scores of the static networks, the plastic networks are less likely to get catastrophically bad scores, when the morphology of the robot is changed slightly. \todo{added line explaining why we cannot expect the same scores for plastic nets under original settings} Note, that since the plastic networks are initialized randomly at the beginning of each new episode, a somewhat lower score for plastic networks under the original training setting is to be expected. This is because we cannot expect the initially random network to perform well in the first few time steps when the learning rules have only had little time to adapt the connections. Static networks, on the other hand, can be optimized to perform well under familiar settings from the very first time step, but suffer from lack of robustness.

Our results indicate that the generalization capabilities tend to improve as the number of rules is reduced. At the end of the \emph{Evolve \& Merge} approach we thus have a model, which has a smaller number of trainable parameters and at the same time  generalizes better. Using a simple clustering algorithm as a way of merging the learning rules, we are able to go from initially having 12,288 rules (corresponding to 61,440 trainable parameters) to having just 384 learning rules (corresponding to 1,920 trainable parameters) while improving robustness, all without increasing the number of generations used for optimization. 

In order to make fair comparisons between the all the models, we did not allow for more generations for the reduced rule sets here. However, it is likely that if we were to permit more optimization time, we could decrease the number of learning rules even further.
The observed robustness cannot just be attributed to a smaller number of trainable parameters, since the smaller static networks that performed at the same level as the plastic networks in the original settings, did not show robust performances in altered settings.

The static network optimized with noisy inputs, on the other hand, had average performances across all novel settings and optimization runs, which were very similar to that of the best plastic network model. Noisy inputs have long been used for data augmentation in supervised tasks, such as speech recognition, to gain more robust models \cite{cui2015data}. Adding noise to inputs has also been used as a way to get robust representations in autoencoders \cite{vincent2008extracting}.
However, using noise in training has been explored to a lesser extend in reinforcement learning-type frameworks \cite{igl2019generalization}. In this paper, the static networks with noisy input provide a strong baseline to compare the reduced plastic networks to in terms of their robustness, and as we can see, the plastic networks achieve similar performances.
However, in order to achieve such results by applying noise to the inputs to static networks, we need to carefully pick the correct amount of noise to apply; too little noise, and the results will be indistinguishable from the plain static approach, and too much noise is likely to hinder progress completely. Using the plastic approach, the parameters will regulate themselves, and with the \emph{Evolve \& Merge} method we have the added benefit of ending up with a smaller number of trainable parameters. 
The similar performances between static networks with noisy inputs and plastic networks are interesting and will have to be explored further in the future. The rather naive approach to using noise of simply adding it on top of the input is unlikely to be a promising method to improve upon to make even more robust models. It is, on the other hand, easier to imagine improvements to the learning rules (see section \ref{future}).

The evolved learning rules used here are inspired by Hebbian learning. However, the model which used only the activity dependent terms of the parameterized rule (the ABC terms), failed to perform well, making it clear that the stability provided by the constant terms are necessary for the locomotion task used here.

The results also showed that starting from a complete rule set, and then merging the rules throughout optimization achieved superior results compared to starting with a small number of rules. This draws parallels to the recent "Lottery Ticket Hypothesis" \cite{frankle2018lottery} that builds upon the notion that a trained neural network can most often be pruned drastically, resulting in a much smaller network that performs just as well as the unpruned network. However, if one was to start training from scratch with a randomly initialized small network of the same size as the pruned network, the training is likely to be much more difficult, and one is unlikely to get the same performance.
The hypothesis states that when initializing a large network, it is likely that one of the combinatorially many sub-networks inside the full network will be "easily trainable" for the given problem; we are more likely to have a "winning ticket" within the random initialization, since we have so many of them. The number of sub-networks - or potential winning tickets -  dwindle rapidly if we decrease the size of the full network.
To the best of our knowledge, the methods used for finding winning tickets \cite{frankle2018lottery, zhou2019deconstructing} have not yet been explored in the case where the optimization method is ES, and much less in the context of indirect encoding. Our results hint that the Lottery Ticket Hypothesis might also hold in the indirect encoding setting that we employ here. Further investigations into when the Lottery Ticket Hypothesis asserts itself in indirect encoding schemes might provide valuable insights for future approaches.

Deciding how to evaluate one’s models on OOD circumstances can seem a bit arbitrary, as countless different changes to a simulated environment can be made. Several previous studies have focused on the ability of a robot show robustness in the face of a severe leg injury \cite{cully2015robots, colas2020scaling, najarro2020}. Here we opted for slight variations on leg lengths instead, partly to highlight how quickly neural network models can break completely. For many of the models, a severe injury is not required in order to put a well-performing model at risk of malfunctioning. Had we chosen a larger range of reductions, it likely that the results would have been tilted to favor the plastic networks more, whereas a smaller range would have had the opposite effect.
The issue of deciding how to evaluate generalization capabilities of a model speaks into a larger discussion of overfitting in artificial agents \cite{zhao2019investigating}. In some sense, the static networks do what we ask them to more so than the plastic ones; when we optimize the networks we implicitly ask them to overfit as much as possible to the problem that we present them. In the approach presented in this paper, we explicitly optimize for one setting, but hope that our models will also be able to perform in different settings. Of course, several already established meta-learning frameworks attempt to create models that learn how to learn to solve new tasks \cite{finn2017model}. However, such approaches also require us to make somewhat arbitrary choices regarding which types of different tasks we should expose our models to during training. For this reason, it is still useful to develop methods that are intrinsically as robust as possible. Such methods might be promising candidates for use in frameworks such as in Model-Agnostic Meta-Learning (MAML) \cite{finn2017model}.

 \subsection{Future Directions} \label{future}
 We showed that it is indeed possible to evolve relatively well-performing models with an increased robustness to morphology changes, while at the same time having fewer trainable parameters; this approach opens up interesting future research directions. 
 
As mentioned in section ~\ref{subsection:heb}, an important concept within Hebbian theory is that of cell assemblies, which after repeated exposure to stimuli can become increasingly correlated and able to perform pattern completion. In this theoretical framework that have been supported by a wide array of evidence\cite{see2018coordinated, miller2014visual, hampson2009neural}(for a review, see \cite{saxena2019towards}), recurrent connections between the neurons are often assumed. In the current study, we have only used simple feedforward networks with two hidden layers. Applying the \emph{Evolve \& Merge} approach to local learning rules for more advanced neural networks with recurrent connections will be an interesting line of research in the future. 
More generally, the \emph{Evolve \& Merge} approach also lend itself well to be combined with methods that evolves the neural architecture of the network that the learning rules apply to. An example of this is the NEAT algorithm \cite{stanley2003evolving, risi2011enhancing, risi2012enhanced}.

It will also be interesting to combine this indirect encoding method with a meta-learning framework such as MAML \cite{finn2017model}. Models with few trainable parameters, controlling a large, expressive ANN might intuitively be an ideal candidate for few-shot learning in the MAML framework.

Further, while local learning rules such as spike-time dependent plasticity are important drivers of change in synapses in the brain, synaptic plasticity is also affected by neuromodulators \cite{dayan2012twenty, feldman2012spike}. Extending the evolved learning rules to be able to take into account reward signals could greatly improve the model's ability to respond to changes in the environment in an adaptive manner, and it is something we look forward to implementing in future studies. Something similar to this has been explored in other approaches to plastic networks \cite{soltoggio2008evolutionary, ellefsen2015neural, bertens2019network, ben2020evolving}, and we expect this to also be a beneficial addition to our approach.  

While we have evolved a set of relatively simple learning rules, our approach could just as well be used in conjuction with more complicated rules, e.g., rules with more parameters \cite{chalmers1991evolution}, or rules that produce non-linear outputs of the inputs \cite{orchard2016evolution,bertens2019network}. With more expressive rules, it might be possible to limit the number of rules and trainable parameters even further.

On a practical note, having few trainable parameters opens up for the possibility of using more sophisticated optimization methods. For example, the Covariance Matrix Adaptation Evolution Strategy \cite{hansen2006cma} has recently been used in several studies to obtain impressive results \cite{ha2018world, tang2020neuroevolution}, but it quickly becomes infeasible to use if the number of trainable parameters is too large.


\section{Conclusion}
If the environment one wishes to deploy an artificial agent in is guaranteed to be identical to the training environment, one might be better off evolving a static network to control the agent. However, if the aim is to have artificial agents act in complex real-world environments, such guarantees cannot be made. The evolution of plastic networks that can better adapt to changes, is therefore an interesting prospect. The results shown in this paper contribute to the development of robust plastic networks. We show that it is simple to achieve a fairly small rule set and that as the number of rules decreases, the robustness of the model increases. This can be achieved with no additional optimization time. We believe that there are several exciting ways to continue this line of research, and we hope that the results presented here will inspire additional studies in evolved plastic artificial neural networks.

\begin{acks}
 This project was funded by a DFF-Research Project1 grant (9131-00042B). 
\end{acks}

\bibliographystyle{ACM-Reference-Format}
\bibliography{refs}


\end{document}